\pdfoutput=1
\documentclass{article}
\usepackage[hang,flushmargin]{footmisc}  
\usepackage[final]{ap_2020}
\usepackage[utf8]{inputenc} 
\usepackage[T1]{fontenc}    
\usepackage{hyperref}       
\usepackage{url}            
\usepackage{booktabs}       
\usepackage{amsfonts}       
\usepackage{nicefrac}       
\usepackage{microtype}      
\usepackage{graphicx}
\usepackage{arydshln}
\usepackage{multirow}
\usepackage{caption}
\usepackage{subcaption}

\title{Emora: An Inquisitive Social Chatbot\\ Who Cares For You}
\author{Sarah E. Finch\footnotemark[1], James D. Finch\footnotemark[1], Ali Ahmadvand, Ingyu (Jason) Choi,\\
\textbf{Xiangjue Dong, Ruixiang Qi, Harshita Sahijwani, Sergey Volokhin,}\\
\textbf{Zihan Wang, Zihao Wang, Jinho D. Choi\footnotemark[2]}\\
Department of Computer Science, Emory University \\
Atlanta, GA 30324 USA \\
\texttt{\{sarah.fillwock, james.darrell.finch, ali.ahmadvand, in.gyu.choi\}} \\
\texttt{\{xiangjue.dong, ruixiang.qi, harshita.jagdish.sahijwani, sergey.volokhin\}} \\
\texttt{\{zihan.wang, zihao.wang2, jinho.choi\}@emory.edu} \\
}

\begin{document}
\maketitle

\renewcommand{\thefootnote}{\fnsymbol{footnote}}
\footnotetext[1]{Team leads. The rest of the student authors are in alphabetical order.}
\footnotetext[2]{Faculty advisor.}
\renewcommand{\thefootnote}{\arabic{footnote}}

\begin{abstract}
Inspired by studies on the overwhelming presence of experience-sharing in human-human conversations, Emora, the social chatbot developed by Emory University, aims to bring such experience-focused interaction to the current field of conversational AI. 
The traditional approach of information-sharing topic handlers is balanced with a focus on opinion-oriented exchanges that Emora delivers, and new conversational abilities are developed that support dialogues that consist of a collaborative understanding and learning process of the partner's life experiences. We present a curated dialogue system that leverages highly expressive natural language templates, powerful intent classification, and ontology resources to provide an engaging and interesting conversational experience to every user.
\end{abstract}

\section{Introduction}

This paper presents Emora, a socialbot capable of engaging and personalized chat on a wide variety of popular topics. 
Our goal is not only to create interesting and fluent dialogue, but to reach beyond information-based chat so that Emora feels more like a caring friend to users than an information desk. Following this objective, Emora supports rich conversation about both popular topics from previous Alexa Prize competitions such as movies, music, and sports, and more personalized topics such as school, work, and hobbies.
To facilitate mixed-initiative conversation, we also provide Emora with her own opinions and personality. 
Our approach engages users through an active curiosity of their opinions and experiences, while relating to them by sharing Emora's own attitudes and profile. 

To ensure conversations with Emora are personalized to the user and highly self-consistent, we take a pipelined approach to Emora's architecture that includes both data-driven and designer-driven components. 
Section~\ref{sec:architecture} describes this architecture to give an overview of Emora's design. 
Section~\ref{sec:nlp} discusses the preprocessing components of our architecture, which provide natural language understanding despite the large variety of expressions that can be present in user utterances. 
Section~\ref{sec:dm} discusses Emora's core decision-making logic for response selection, which takes a state machine-based approach that is augmented by components designed for flexible topic transitioning. 
Section~\ref{sec:topics} outlines the topics Emora supports and highlights differences in challenges and approach present in each topic. Finally, Section~\ref{sec:analysis} gives an analysis of our social bot. We find that our direction of creating a more socially-driven conversation focusing on personal information and experiences is highly successful, yielding an average rating of 3.61 over the last seven days of the semifinal.

\section{Related Work}

Previous work on human-computer dialogue systems has focused on two main tracks, task-oriented dialogue and open-domain chat. 
Task-oriented systems focus on completing a predefined objective, such as booking tickets or scheduling a restaurant reservation \citep{BordesW16,rajendran-etal-2018-learning}. Following the format of the Alexa Prize Competition, our social bot is more in line with previous work in open domain chat. 

Open-domain chat work broadly consists of approaches involving both retrieval models like that of \cite{gu2019dually}, \cite{lu2019constructing}, and \cite{yuan2019multi} and generative models such as that in \cite{luo2018auto},  \cite{parthasarathi2018extending}, and \cite{adiwardana2020towards}. While these data-driven neural approaches are highly scalable to virtually any conversation domain, their reliance on large sets of training data makes them difficult to control. Inspired by work indicating that most human-human conversation involves discussions of personal information, opinions, and experiences \citep{mitsuda:19}, we choose a pipelined approach to our dialogue system that includes handcrafted and rule-based processing components in addition to neural models and databases. This allows us to tightly control the content and dialogue strategies of Emora, giving her a rich and self-consistent personality and user experience. 

Reliance on rule-based content to create highly personalized social bots is not a new approach in previous Alexa Prize competitions \citep{alexaprize}. Despite this, social bots from previous competitions rely heavily on information-oriented conversation. A common strategy is to recognize an entity in the user utterance, and reply with facts or trivia about the entity. When user and system opinions are discussed, they usually regard attitudes towards third-party entities rather than discussing entities that are regularly involved in the user's daily life. We aim to distinguish Emora's conversational focus to be more in line with content that appears in human conversation, where personal circumstances and experiences take center stage. In Section \ref{sec:analysis_personal}, we present results that this personalized approach to chat is highly successful.
\vspace{-2ex}
\section{Emora System Architecture}
\label{sec:architecture}
\vspace{-1ex}

Emora is implemented using the Amazon Conversational Bot Toolkit (CoBot), which provides an automated and low-effort deployment infrastructure for socialbot components with appropriate auto-scaling policies enabled \citep{khatri2018advancing}. 
CoBot utilizes AWS EC2 and ECS for the deployment of components and DynamoDB to store user information and conversation records. Within the Cobot framework, a socialbot interfaces with a user through an AWS Lambda function where user utterances trigger events to which the socialbot responds. These sequences of events occur in a parallelized manner across multiple users.

\begin{figure*}[!htbp]
\centering
\includegraphics[width=\textwidth]{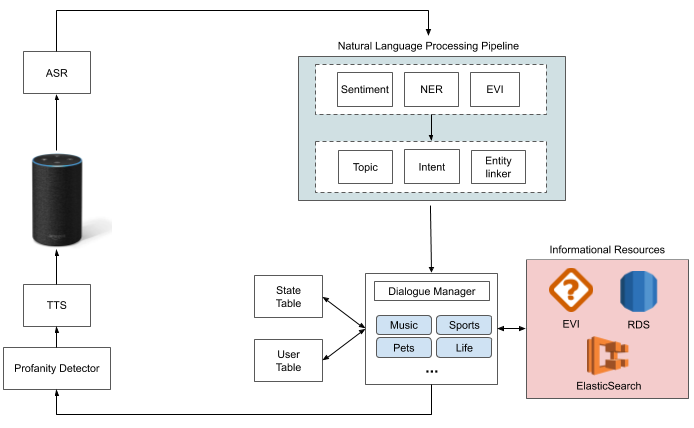}
\caption{Overall system architecture of the Emora socialbot from reception of user utterance to production of system response.}
\label{fig:system}
\end{figure*}

Figure \ref{fig:system} illustrates Emora's overall system architecture. When a user connects to Emora through the Alexa skill, \textit{Let's chat}, their utterance is transcribed by the built-in Automatic Speech Recognizer. We receive the most probable text transcription as well as the list of candidate hypotheses. This text transcription is sent to our Natural Language Processing (NLP) Pipeline, which performs feature extraction and text classification tasks whose results can be utilized by later components in the architecture. The NLP Pipeline is split into two consecutive rounds, where the components in each round run in parallel with one another. The boxes with dotted lines denote the two rounds in Figure \ref{fig:system}. 

Once the NLP Pipeline has completed, the Dialogue Manager is invoked. Based on the current dialogue state, the extracted features, and the user's current utterance, the Dialogue Manager selects the most appropriate system response, which will be described in depth in Section \ref{sec:dm}. The entry point to each topic handler is housed in the Dialogue Manager's control logic, although the topic handlers themselves are free to use any features and external resources necessary. Within several of our topic components, additional Amazon services are used for efficient storage and retrieval of topic-specific information. These services include Amazon Relational Database Service and Amazon Elasticsearch, and will be described in more detail in Section \ref{sec:topics}. Once a system response is selected, the provided Profanity Detector Service from Amazon determines whether there is any sensitive content with the response; otherwise, the system response is returned to the Lambda function from which it is spoken to the user. If there is sensitive material detected, we return a fallback response instead, which is randomly selected from a collection of generic responses, some of which are used only as a fallback to specific components. At the end of each turn, turn-based dialogue state information and learned user attributes are stored into the State and User DynamoDB tables, respectively, and the system response is returned to the Lambda function for speech synthesis.

\section{Natural Language Processing Pipeline}
\label{sec:nlp}

The initial processing of a user utterance is the Natural Language Processing Pipeline, which aims to extract basic text features that are useful to later conversational processing. The output of each of the components described in this section are available to all components, including the dialogue manager and the topic handlers.

\paragraph{\bf Sentiment}
We use the Valence Aware Dictionary and sEntiment Reasoner (VADER) to estimate a sentiment score for each received user utterance \citep{hutto:14}. VADER is specifically tuned for social media and microblog, which is applicable to the casual, short nature of dialogue utterances, and provides low latency due to the rule-based lexicon formulation of the VADER model.

\paragraph{\bf Named Entity Recognition}
Named entities are identified within user utterances using the SpaCy named entity recognizer (NER).\footnote{\url{https://spacy.io/usage/linguistic-features\#named-entities}} SpaCy NER is a statistical model trained on the OntoNotes 5 corpus which recognizes 18 classes of named entities, including \textit{person}, \textit{organization}, and \textit{location}.

\paragraph{\bf EVI}
The Amazon-provided EVI question-answering service is used in a similar manner to more traditional NLP components. The user utterance is sent to EVI and the returned response is stored as a feature that can be accessed by any later components.

\paragraph{\bf Entity Recognition and Linking}
We use a comprehensive index of entity names and their various surface forms to recognize entities. We maintain an index of more than 3.7 million entities and their types. It is regularly updated with new entities that our domain-specific components can handle, e.g. movie names from RottenTomatoes, and people and organizations from the Washington Post API. We also indexed all the entities present in the DBPedia snapshot from 2016 \footnote{\url{https://wiki.dbpedia.org/develop/datasets/dbpedia-version-2016-10}} to improve coverage of less recent entities. Given an utterance, we match all ngrams in it against our index. 

\paragraph{\bf Topic and Intent Classification}

In open-domain conversational agents, topic and dialogue intent classification can be treated as a text classification problem \citep{rf:emerson, ahmadvand2018emory}. However, compared to general text classification, utterance classification poses a greater challenge due to four main factors: 1) the tendency of human utterances to be short; 2) errors in Automatic Speech Recognition (ASR); 3) users’ frequently mentioned out-of-vocabulary words and entities; and 4) lack of available labeled open-domain human-machine conversation data \citep{rf:CDAC, rf:concet}. 

\begin{figure*}[htbp]
\includegraphics[width=\textwidth]{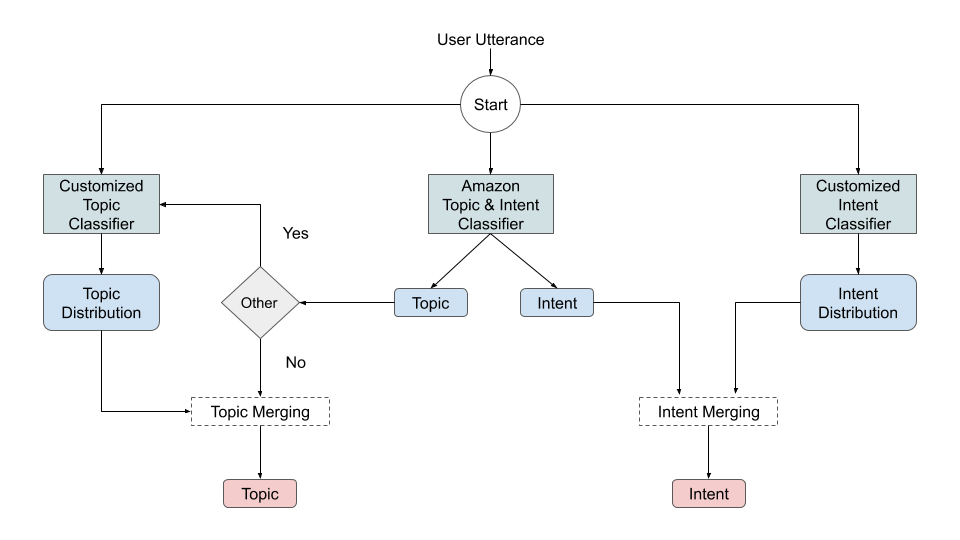}
\caption{Flowchart of the Mixture of Experts (MoE) Model for topic and intent classification. The green rectangles are the three models utilized in the MoE, the blue rectangles are intermediate processing outputs, and the red rectangles are the final outputs.}
\label{fig:TopicIntent}
\end{figure*}

Moreover, natural conversations entail utterances which are dependent on the context, thus making it impossible to classify the topic and intent without considering the preceding utterances. For example, when a user uses the expression, ``Oh, yeah'' it can be interpreted as one of several options, such as "Accept-Agree" or "Topic-Switch." To help address these problems, we developed a contextual-aware topic and dialogue intent classification model for open-domain conversational agents.

To identify both {\em topics} and {\em dialogue intents}, we developed a Mixture of Experts Model that is composed of the classifier provided by Amazon as well as our own customized topic and dialogue intent classifiers. These customized models are deep learning models which were developed for last year's competition. The details and performance evaluation of these customized classifiers are described in \cite{ahmadvand2018emory}. These customized models are trained on our internal conversation dataset which was collected and manually annotated from last year's competition. 

Figure \ref{fig:TopicIntent} shows the general flow of Emora's topic and dialogue intent classifiers, where the Amazon classifier plays a principal role in the classification. The outputted topic and intent labels from the customized classifier are used to update the label probability distribution produced by the Amazon classifier.

The customized classifiers aim to improve coverage of the Amazon-provided classifier. For example, we train our topic classifier on new topics such as ``Food-Drinks'' and ``Travel-Geo'', and we train our intent classifier on new dialogue intents such as ``Yes-Answers.''  In addition, the customized classifiers aim to improve the quality of some specific classes to better fit our use cases. For instance, we observed that Amazon's intent classifier often considers user disagreements as ``Topic-Switching'', even when this is not the most accurate label of the current dialogue situation. To alleviate this problem, we added another intent called ``Reject'' to cover the situations where the user was disagreeing with the last system response but did not intend to end the current topic being discussed.

\vspace{-2ex}
\section{Dialogue Manager}
\label{sec:dm}

\begin{figure*}[!htbp]
\centering
\includegraphics[width=\columnwidth]{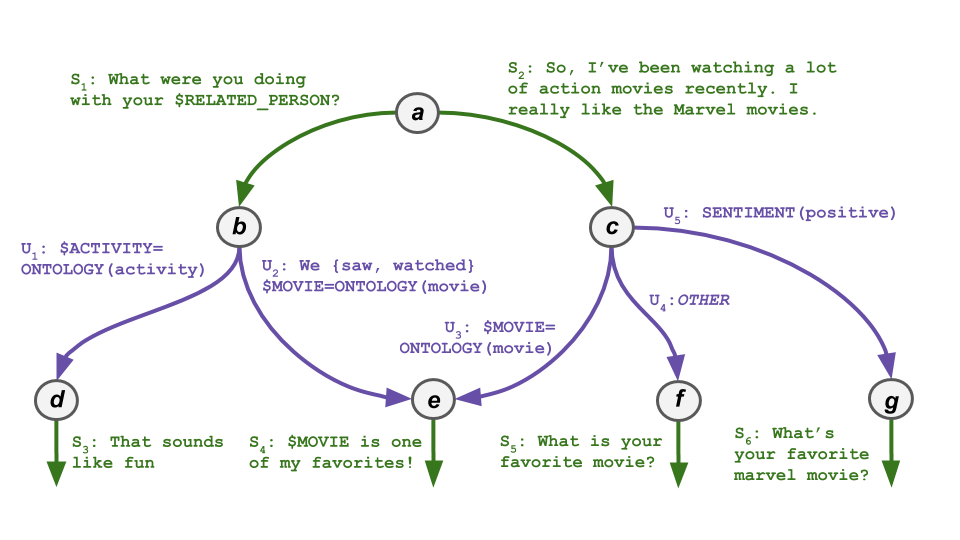}
\caption{A simplified example state machine used to manage Emora's dialogue. System transitions are marked $S_1$ to $S_6$ in green, and user transitions are marked $U_1$ to $U_5$ in purple. User transitions labeled with "ONTOLOGY" reference nodes in the ontology for NLU, transition $U_2$ uses rule-based pattern matching, and transition $U_5$ uses a sentiment classifier. Capitalized tokens preceded with "\$" denote variables. Figure adapted from \cite{finchsigdial}.} 
\label{fig:statemachine}
\end{figure*}

The dialogue manager is responsible for selecting the system response given the dialogue context. Our dialogue manager is based on a state machine, where states efficiently represent the dialogue context and transitions represent either system or user turns. The state machine models turn taking by alternating between states dedicated to either the system or user turn. 
For each user or system turn, outgoing transitions from the current state represent possible dialogue actions of the speaker. Figure~\ref{fig:statemachine} shows an example, where user transitions are in green and system transitions in purple.

To perform natural language understanding (NLU), each user transition in the state machine is labeled with a reference to a classifier that indicates whether a given user utterance matches the transition. We use a combination of rule-based matching and neural models for our transition classifiers, such as the topic and intent classifiers discussed in Section \ref{sec:nlp}. Additionally, many user transitions are augmented by tagging models that extract useful entities or other information from the user utterance. Our taggers are mainly implemented using rule based methods that leverage our ontology to optimize precision, but our high-recall entity linker is also used for this purpose when named entities are the extraction target. Tagged information from the user utterance is stored in a variable table, which is referenced by future transitions, both to open up new valid transition pathways and to fill slots during natural language generation.

Our system's natural language generation is primarily done using handcrafted templates. We find that handcrafted templates are the most reliable way to produce natural-sounding responses from the system. These templates contain slots referencing variables from the variable table, which can be used to personalize the conversation using information the user provides. As an example, state $a$ in Figure \ref{fig:statemachine} is a system state with multiple system responses. The response represented by transition $S_1$ references a variable $\$RELATED\_PERSON$, and will only be selected as a valid transition if that variable has been set by a previous transition. This results in the transition being taken when information about the user is known, such as if Emora knows the user was spending time with their friend. In this case, Emora can respond "What were you doing with your friend?", signaling comprehension to the user and creating a more personalized interaction. If this information was not learned from the user in previous turns, transition $S_1$ would be invalid and transition $S_2$ would dictate the system response as a default option.

\subsection{Ontology}

To improve the generalizability of our NLU classifiers, our dialogue manager is equipped with an ontology. We populate this ontology from a variety of sources, including WordNet \citep{fellbaum2012wordnet} and Wikipedia lists. Especially when using high-precision NLU with pattern matching, the ontology allows Emora to account for many synonyms, hyponyms, and idioms of user expressions and recognize a variety of entity types.

\begin{figure}
\centering
\begin{subfigure}{.5\textwidth}
  \centering
  \includegraphics[width=\linewidth]{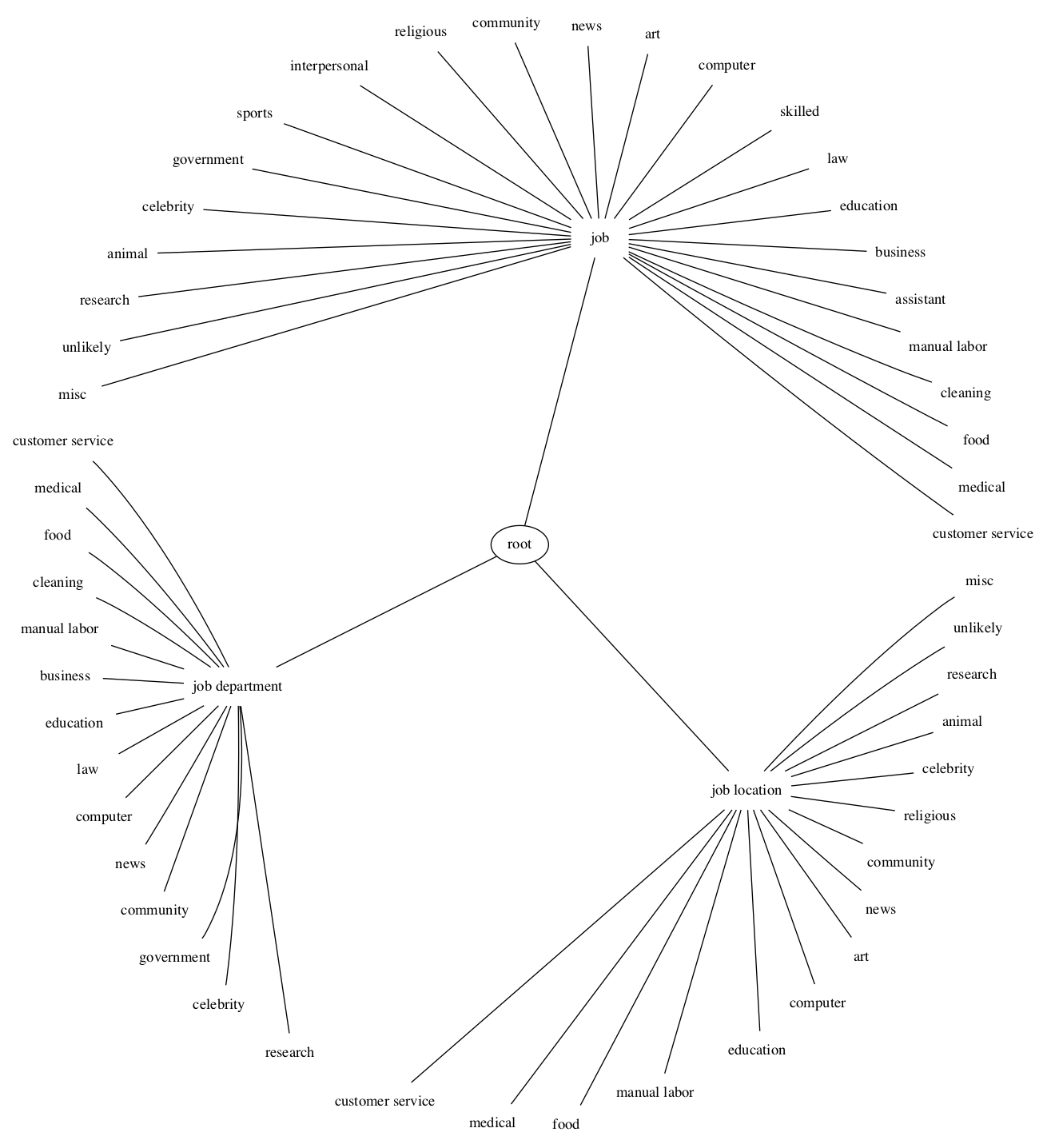}
  \caption{Job-related terms}
  \label{fig:ontjob}
\end{subfigure}%
\begin{subfigure}{.5\textwidth}
  \centering
  \includegraphics[width=\linewidth]{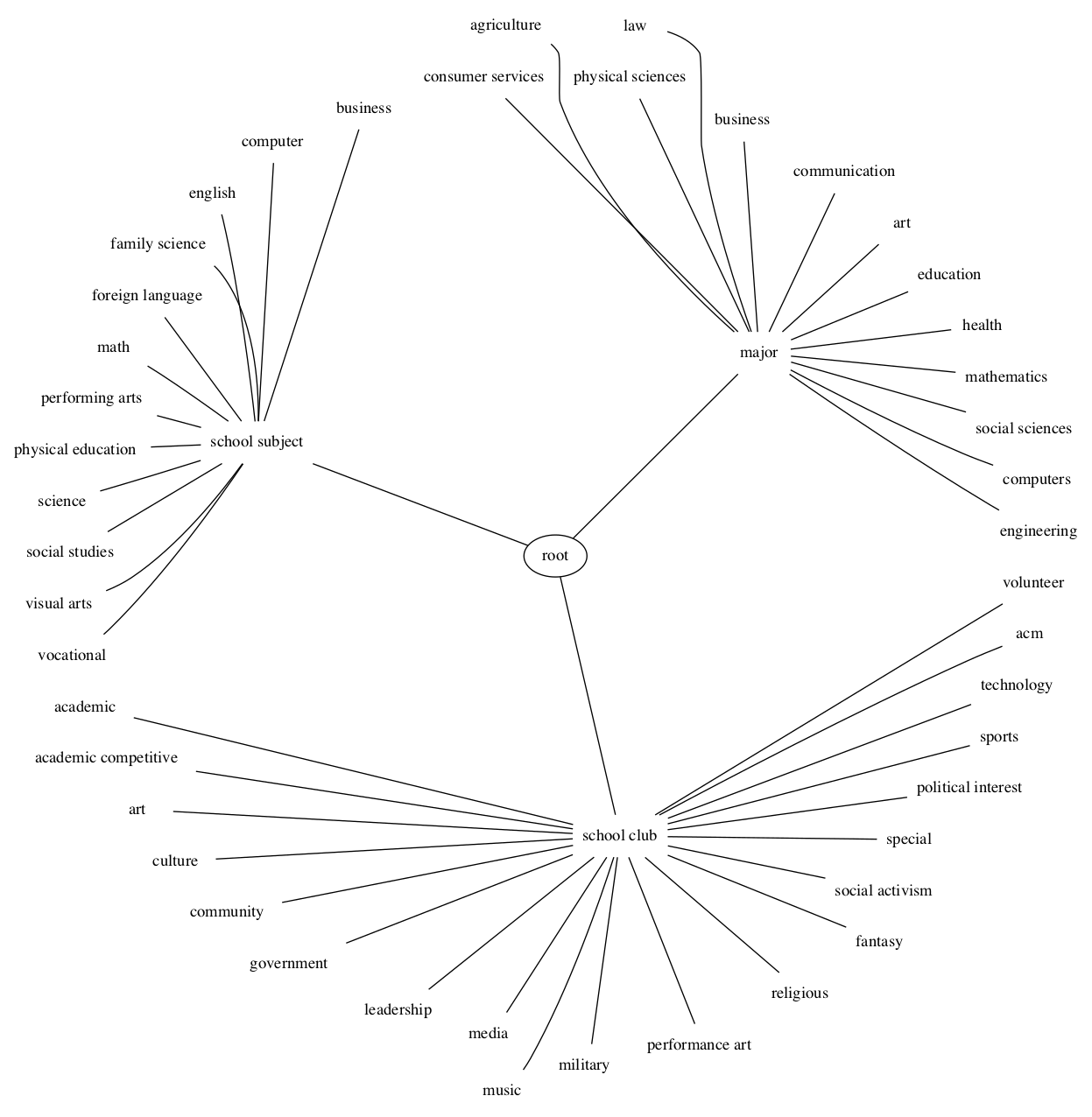}
  \caption{School-related terms}
  \label{fig:ontschool}
\end{subfigure}
\caption{Two segments of the ontology used for NLU in the Emora socialbot.}
\label{fig:ontology}
\end{figure}

Figure \ref{fig:ontology} illustrates the substructure of two segments of the ontology. For the sake of figure clarity, the terminal lexical strings are not shown. Although our figure only shows six top-level nodes, our ontology also includes:
\begin{itemize}
    \item life events (e.g. illness, birth, romantic event, etc.)
    \item related persons (e.g. family, partner, etc.)
    \item states of being (e.g. injury, positive state, negative state, etc.)
    \item locations (e.g. cities, countries, etc.)
    \item activities (e.g. chore, errand, exercise, etc.)
    \item life stages (e.g. child, teenager, adult, student, etc.)
    \item status (e.g. employment status, relationship status, education status, etc.)
    \item political affiliation
    \item emotions
    \item adjectives
    \item animals
    \item names
    \item personality traits
\end{itemize}

\subsection{Inter-Topic Transitions}
\label{sec:transitions}
Although state machine-based dialogue managers have the advantage of an efficient context representation and offer a high degree of developer control over conversation flow, they have little support for flexible transitions between topics. A common case is that the user requests to switch to a new topic, which is possible at virtually any point in the conversation. For a state machine to represent this possibility, an additional transition would have to be added for each combination of $(user\_state, topic\_start)$ state pair. Therefore, our dialogue manager is augmented by a global transition table, which includes states that can be transitioned to from anywhere in the state machine, along with their corresponding NLU classifiers. 

To further improve the fluency of topic switches in conversation, our dialogue manager also includes a state stack. The purpose of the stack is to record states that can be returned to after transitioning into a different topic as a sub-dialogue. Any transition in the state machine can optionally push a state onto the stack, or choose to return to the state on the top of the stack. For example, a conversation in the COVID-19 topic may take a global transition into a subtopic about the user's experiences with remote learning during the pandemic. This subtopic is simply implemented as an extension of the COVID-19 state machine, but is accessed flexibly via global transitioning. Since it is natural to return to the higher-level COVID-19 discussion once this subtopic is complete, entering this subtopic would push the previous COVID-19 state onto the stack and return to that state after the subtopic is over. In the next section, we describe an example conversation Emora is capable of, which includes usage of the stack to improve topic transitioning fluency.
\section{Conversation Flow} 
\label{sec:flow}

Following our goal to design Emora as a chatbot capable of engaging in human-like social aspects of conversation, we design our system to reach beyond opinion-oriented conversation and support personal conversation. We distinguish personal conversation from opinion-oriented conversation based on whether speakers are sharing thoughts and attitudes towards entities that are regularly present and impactful to their daily lives. For example, stating an opinion about a movie would be regarded as opinionated but not personal, whereas a speaker expressing an opinion towards a class they are enrolled in would be considered personal, since the class has a direct and substantial involvement in the speaker's own life. Although we recognize that this distinction exists along a continuum, we find it helpful to explicitly focus on driving a more personal conversation to improve the human-like qualities of Emora. This includes developing a profile for Emora's own personality and experiences. Details of this profile are given in descriptions of our various topic handler components in Section \ref{sec:topics}.

Table \ref{tab:example} presents a simulated conversation with our socialbot. The given conversation is an exemplification of the personal-experience-focused dialogue strategy that is the main novelty of our approach.  It provides the general interaction style of sharing and receiving personal experiences with the user in a manner that balances displaying an interest in the user's experiences through question-asking and giving open-ended statements that allow the user the freedom to respond as they wish. 

The \textit{Variable Table} column denotes the learned user attribute information from specific utterances, e.g., the user's status as a student based on their mention of online courses in \texttt{U2}. Learning such information allows for Emora to select topics of conversation at later points that connect directly to the user's interests and life, as shown in \texttt{E9(b)}, which deepens the user's feeling of being understood.

The \textit{Stack} column shows the progression of stack operations that occurs during the course of the example conversation. As explained in Section \ref{sec:transitions}, adding and removing dialogue states on the stack are specified on a transition level. For instance, the user transition satisfied by \texttt{U1} has been identified as a subtopic initiation and, thus, the original topic's next state (\texttt{covid\_sympathy}) is pushed onto the stack in order to preserve its progress and allow the option of returning to it at a later point. Turns \texttt{U1} to \texttt{E3(a)} result from the remote learning subtopic. Upon completion of the subtopic at Turn \texttt{E3(a)}, the saved dialogue state on the stack is considered and ultimately chosen as the next conversational topic. This procedure of selectively adding dialogue states as future points to return to in the conversation after completing subtopics increases the overall feeling of coherency in the conversation and allows Emora to utilize conversational memory to drive the conversation forward. 

Although not explicitly illustrated in the example, the stack operations are not constrained to having a single dialogue state on the stack at a time. At any point at which a subtopic is initiated, even if already conversing within a different subtopic, the stack is updated to maintain the history of unfinished topics. Then, the option at later points in the conversation becomes which dialogue state on the stack to return to. A life counter is included for all dialogue states in order to allow for case-by-case optimization of the user experience, where a dialogue state on the stack is no longer a valid option to return to after a specified number of turns have elapsed. However, the general strategy is to return to previous dialogue topics in the traditional stack order (Last In, First Out) to preserve conversational coherence.

\begin{table}[htbp!]
\centering\resizebox{\columnwidth}{!}{
\begin{tabular}{c|l|c|c}
\multicolumn{1}{c|}{\textbf{ID}} & \multicolumn{1}{c|}{\bf Utterance} & \textbf{Variable Table}       & \textbf{Stack}          \\
\hline\hline
\multirow{2}{*}{\texttt{E1}}
 & I've been hearing a lot from people about how strange it is & & \multirow{2}{*}{{[} {]}} \\
 & to live with the corona virus. Has your life changed much? & & \\
\hdashline[1pt/1pt]

\multirow{2}{*}{\texttt{U1}}
 & \multirow{2}{*}{Yeah my school has online courses now} & \tt student=True & \multirow{2}{*}{{[}\texttt{CoS}{]}} \\
 & & \tt remote=True & \\
\hline

\texttt{E2} & Oh, are you liking your online classes?        &   & {[}\texttt{CoS}{]}   \\
\hdashline[1pt/1pt]
\texttt{U2}  & Not really         &   & {[}\texttt{CoS}{]}   \\
\hline

\multirow{2}{*}{\texttt{E3(a)}}
 & Oh yeah, it is hit and miss for a lot of people. & & \multirow{2}{*}{{[}\texttt{CoS}{]}} \\
 & Definitely just try to do the best you can. & & \\
\hdashline[1pt/1pt]

\multirow{3}{*}{\texttt{E3(b)}}
 & Actually, I have seen something pretty uplifting recently. & & \multirow{3}{*}{{[} {]}} \\
 & Have you seen those videos where zoos are letting some & & \\ 
 & of their animals out of their cages to roam around? & & \\
\hdashline[1pt/1pt]

\texttt{U3} & No  &   & {[} {]}   \\
\hline

\multirow{2}{*}{\texttt{E4(a)}}
 & My favorite one was the penguins at the Chicago zoo, & & \multirow{2}{*}{{[} {]}} \\
 & you should check it out! & & \\
\hdashline[1pt/1pt]
\multirow{4}{*}{\texttt{E4(b)}}
 & Anyway, the good news is that this virus won't last forever  &  & \multirow{4}{*}{{[} {]}} \\ 
 & and people are taking steps in the right direction to lower & & \\ 
 & its impact. I know things might seem weird right now, but & & \\
 & just do the best that you can and stay positive. & & \\
\hdashline[1pt/1pt]

\multirow{2}{*}{\texttt{U4}}
 & I will try to but its pretty stressful, i am having a hard time & \multirow{2}{*}{\tt need\_grocery=True} & \multirow{2}{*}{{[}\texttt{CoE}{]}} \\
 & buying groceries & & \\
\hline

\multirow{4}{*}{\texttt{E5}}
 & Yeah, that's tough. None of my friends can find necessities & & \multirow{4}{*}{{[}\texttt{CoE}{]}} \\
 & like toilet paper in stores anywhere. Honestly, who would & & \\ 
 & have ever imagined there would come a time when toilet & & \\ 
 & paper was such a sought after item? & & \\
\hdashline[1pt/1pt]

\texttt{U6}  & It is crazy        &   & {[}\texttt{CoE}{]}        \\
\hline

\texttt{E6(a)} & For sure.          &   & {[}\texttt{CoE}{]}        \\
\hdashline[1pt/1pt]

\texttt{E6(b)} & So, what did you get up to today?   &   & {[} {]}   \\
\hdashline[1pt/1pt]

\texttt{U6}  & I called my mom   & \tt activity=talk$\_$to$\_$mom & {[} {]}   \\
\hline

\multirow{2}{*}{\texttt{E7}}
 & It is good to hear that you are keeping in touch with the &  & \multirow{2}{*}{{[} {]}}  \\
 & people in your life. Are you close to your mom? & & \\
\hdashline[1pt/1pt]

\texttt{U7} & We are pretty close &  & {[} {]}\\  
\hline

\multirow{2}{*}{\texttt{E8(a)}}
 & Wow, you sound like you are really close to your family.  & \multirow{2}{*}{\tt close$\_$to$\_$mom=True}  & \multirow{2}{*}{{[} {]}} \\
 & I am glad to hear that. & & \\
\hdashline[1pt/1pt]

\multirow{2}{*}{\texttt{E8(b)}}
 & So, I remember you saying something about being a student.  &  & \multirow{2}{*}{{[} {]}} \\
 & I am actually a student too. Do you like school? & & \\
\hdashline[1pt/1pt]

\texttt{U8} & Yes I do & \tt likes$\_$school=True & {[} {]}\\

\end{tabular}}
\vspace{0.5em}
\caption{Simulated interaction with Emora, modeled on interactions observed from real users. User inputs are denoted as \texttt{U\#}. System responses involving multiple state transitions are denoted on separated contiguous lines as \texttt{E\#[a-b]}. The third column indicates the user profiling information extracted from a given user utterance. The fourth column displays the contents of the stack at each point in the conversation (\texttt{CoS}: \texttt{\texttt{covid\_sympathy}}, \texttt{CoE}: \texttt{\texttt{covid\_end}}).}
\label{tab:example}
\end{table}









\section{Topic Handlers}
\label{sec:topics}
In this section, we present the design and implementation details of our various topic handlers. 

\paragraph{\bf COVID-19}
The \textit{COVID-19} component was developed in response to the rapidly progressing state of the world due to the emergence of the pandemic. We developed two main different conversations about COVID-19, a comparison of which is presented in Section \ref{sec:corona_result}. One of our conversational tracks was designed to inform users about useful facts and up-to-date information of the pandemic, including confirmed cases and mortality rate of specific regions. The detailed statistics were regularly updated based on daily reports from Center for Systems Science and Engineering (CSSE) at Johns Hopkins University\footnote{\url{https://github.com/CSSEGISandData/COVID-19}}. Our second conversational track took a more opinion-oriented approach to discussing the ongoing events of COVID-19, focusing on how the user was being affected by the changes that the virus was causing on society as a whole.



\paragraph{\bf General Activities}
The purpose of the \textit{General Activities} component was to elicit the user's daily activities, motivated by the assumption that the user's daily activities are a strong indicator of topics that relate to the user's personal interests. This component was developed in two stages: an exploratory phase and an exploitation phase. Within the exploratory phase, we deployed a small version of the component that demonstrated little natural language understanding of the user's daily activity responses, instead providing a generic acknowledgement and randomly selecting the next topic. Over a month-long period, we collected the daily activity responses, ranked them according to frequency, and manually curated responses to the most frequently occurring types of daily activities, which included work, school, errands, chores, and video games. For the exploitation phase, we then deployed this updated version of the \textit{General Activities} component, in which these small developed interactions on daily activity topics act as smooth transition between the opening of the overall conversation and our other topic handlers.

\paragraph{\bf Holiday}
Over the course of the Quarterfinals and Semifinals period of the competition, we developed brief interactions for various popular holidays, including Valentine's Day, April Fool's Day, and Easter. These interactions were designed as a temporary opening to our conversations, and focused on eliciting the user's plans for the holiday. By design, they lasted no more than 5 turns and transitioned into the core components of our dialogue system upon completion.

\paragraph{\bf Life}
\label{sec:life}

The \textit{Life} component is composed of 6 subcomponents, all of which have the underlying theme of being directly related to a user's personal life experiences. The subcomponents include:
\begin{itemize}
    \item Children
    \item Home
    \item Relationships
    \item Siblings
    \item School
    \item Work
\end{itemize}

The \textit{Life} component focuses on understanding the user's feelings and thoughts on the topics that dominate daily life. Each subtopic in the \textit{Life} component is individually implemented, but the learnable user attributes and conversational paths available in each subtopic are tightly connected to other \textit{Life} subtopics in order to emulate human discussions as naturally as possible.

In addition, this component focuses heavily on expressing Emora's own background information, allowing for the user to interact with, question, and understand Emora's personality and experiences. For example, we give Emora the profile of a student who loves outdoor activities, has a dog, and regularly socializes with friends. We find that providing Emora with this personal information facilitates interest and engagement by allowing the user to reciprocate many of the questions that Emora asks about the user's experiences.

\paragraph{\bf Movies}
The \textit{Movies} component is a topic handler adapted from the Iris bot, which was a part of Alexa Prize 2018 and is described in detail in \cite{ahmadvand2018emory}. Similar to last year, the movie component focuses both on opinion-oriented questions to engage the user and detect their preferences, and on recommending new movies to the user based on these learned preferences.


The movie data used in this component comes from RottenTomatoes and includes various metadata, including genre, actors, plot, ratings, and more. The data is then indexed into 2 separate Elasticsearch indexes based on whether the movie is currently in theaters. Movie data was crawled continuously throughout the day on a dedicated EC2 instance and the indexes were updated nightly. In addition, Python's {\em imdbpy} library was used to query IMDB for trivia facts about movies in order to add variety to the conversation with relevant and interesting remarks.

\paragraph{\bf Music}
The \textit{Music} component's design is an enhanced version of the music component from Iris bot of Alexa Prize 2018, and implementation details can be found in \cite{ahmadvand2018emory}. Just as the \textit{Movies} component does, the \textit{Music} component also interacts with the user to gather indicators of interest for music genres and artists using dynamic templatized questions, and then either recommends similar artists or genres or provides general information of interest. The data comes from various sources, using multiple open Web APIs to gather the data, including Billboards, SeatGeek, Spotify, and lastFM. The data is stored in several DynamoDB databases.

\paragraph{\bf News}
The \textit{News} component is responsible for responding to queries about current events. It is adapted from Iris bot of Alexa Prize 2018 \citep{ahmadvand2018emory}. Given some news topic of interest to the user, we perform BM25 search over the title, category, and body fields of each news article in our news index in order to obtain a set of relevant results, from which we select one and give the user a short summary of it. The component aims to refine further presented news results by prompting the user for additional categories of interest.

We maintain an index of news articles using Amazon Elasticsearch. These news articles are crawled from various sources including The Washington Post API, News API\footnote{\url{https://newsapi.org/}}, New York Times Article Search API\footnote{\url{https://developer.nytimes.com/docs/articlesearch-product/1/overview}}, Reuters newswire and RSS feeds.

\paragraph{\bf Pets}
The \textit{Pets} component carries on a conversation with the user about any pets that they may have. It has separate talking points depending on what type of pet the user has, where these talking points are tailored to the common characteristics of the animal, including cats liking to hide and a dog's ability to learn tricks. Regardless of the kind of pet, Emora will ask the user about their favorite thing about their pet and who their pet seems the most attached to in their family. Throughout the \textit{Pets} conversation, Emora will reciprocally share experiences she has had with her own pet dog in an effort of establishing shared experiences and generating interest with the user. In addition, to add variety to the \textit{Pets} interactions, we scraped information on animal movies and animal jokes from the web, which were integrated as information that Emora produces at various points during the \textit{Pets} conversation.

\paragraph{\bf Sports}
The \textit{Sports} component is capable of talking about the four most popular sports according to user logs, which are basketball, football, soccer and baseball. Sports conversations are driven using a database associated state machine, carefully-designed conversation flows, and recent event updates.

Our \textit{Sports} database was constructed on Amazon Relational Database Service (AmazonRDS) and is regularly updated to contain real-time information such as recent sports games, player statistics and team statistics that are scraped from `https://www.sports-reference.com/'. Based on this database, Emora is capable of talking about most recent games, popular players and teams. For example, it talks about recent interesting NBA games and supports sharing opinions on players' performance in those games. Furthermore, the \textit{Sports} component also shares its own experience to users, which makes it more human-like. For example, it talks about its own basketball playing style and its experience of going to live sports games. Finally, sports is capable of talking about current events such as the NBA shutdown due to COVID-19 and the untimely passing of Kobe Bryant. 


\paragraph{\bf Travel}
The main task of the \textit{Travel} component is to interact with users on certain aspects of cities that users care about the most when traveling, such as tourist attractions, culture, food, events and festivals. To initiate the conversation, the \textit{Travel} component proposes to talk about a city which is randomly generated from a predetermined list of cities that have been designated as cities that Emora is interested in. If the user has not been to the city before, then Emora will continue the conversations like a real friend does, proposing potential plans for traveling to this city. If, however, the user has been to the proposed city, Emora will show an interest in the user's opinions and experiences to give the feeling of being a caring friend. By mixing socially-oriented user experience design with our data on relevant city information, users and Emora create engaging conversations on traveling together.

\paragraph{\bf Teleportation}
\textit{Teleportation} is the smallest permanent topic handler that was developed and was used as an attention grabbing intermediate point of conversation between our more extensive topics. The goal of this topic handler is to create a more niche conversation as an alternative to popular topics. \textit{Teleportation} presents the idea of our socialbot being interested in science fiction, from which a fascination with teleportation has developed. The conversation revolves around a discussion of the user's and Emora's opinion on the likelihood of teleportation in the future and its advantages and disadvantages. Emora initiating this topic is meant to give the sense that she has her own agency and interests, since few users will explicitly request to talk about teleportation themselves.

\paragraph{\bf Virtual Reality}
Our \textit{Virtual Reality} component favors depth over breadth of technology-related topics. It is composed of an experience-oriented discussion on the virtual reality and augmented reality technologies of today, relating to the user's own personal interaction with either technology as well as eliciting the user's opinions on the utility of the technology to the general population.

\paragraph{\bf Video Games}
Our \textit{Videogame} component encodes discussions of twenty popular video games, so that Emora is able to carry on a unique game-specific conversation with user's who identify one of the twenty as their favorite game. These popular video games include: Pokemon, Super Smash Brothers, Skyrim, and Animal Crossing. However, given a non-recognized favorite videogame, Emora interacts with the user in a learning-oriented manner, expressing interest in learning specific characteristics of the unknown game, including its genre and its game mode. In addition, regardless of favorite game, the \textit{Videogame} component relies heavily on a meta gaming conversation flow, which focuses on discussing the future of the industry with the user and how to practice gaming in a healthy way.

\section{Results and Analysis}
\label{sec:analysis}

\begin{figure*}[!htbp]
\centering
\includegraphics[width=0.9\columnwidth]{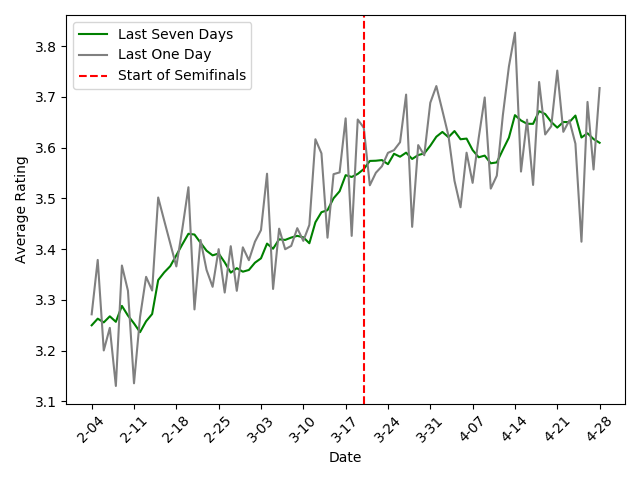}
\caption{The daily average user rating and the average seven day user rating received during the Quarterfinals and Semifinals.}
\label{fig:ratings}
\end{figure*}

During the course of the Alexa Prize Competition, we received daily feedback as to the performance of our socialbot based on user ratings. Figure \ref{fig:ratings} shows the daily and last seven day average user ratings that we received during the course of the Quarterfinals and Semifinals periods. There is a definite positive trend, providing empirical support that the various changes and system updates we have released over this time period have succeeded in improving the experience of having a conversation with our socialbot.

\begin{figure*}[!htbp]
\centering
\includegraphics[width=0.9\columnwidth]{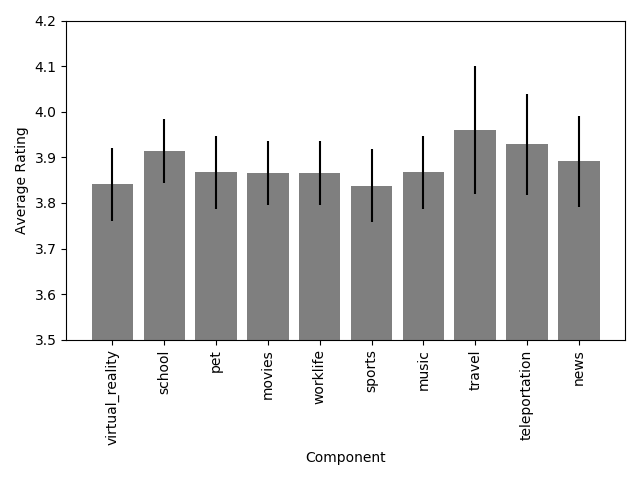}
\caption{The average rating for conversations containing each topic component over the last two weeks of the Semifinals.}
\label{fig:components}
\end{figure*}

\subsection{Component-Level Analysis}
\label{sec:analysis_personal}

Although the overall conversation rating provides a holistic view of the performance of our system, we also conducted a breakdown of the satisfaction that users indicated with specific components of our system. To approximate such a satisfaction measurement, we took the set of all conversations that contain a specific component and calculated the average rating over that set. Given that each of our components had approximately equal distribution within a conversation, this method for calculating the average rating is unaffected by position in the conversation or conversation length. This component-wise breakdown of the performance of our system is shown in Figure \ref{fig:components}. The average rating of each component is composed of all conversations in which the component was invoked over the final two weeks of Semifinals period. The bars denote the 80\% confidence interval. 

Although the system-level component-wise analysis suggests which components performed the best according to the users, we also conducted more detailed investigation into the differences between some of our components' approaches and present those results in the following sections. 

\subsection{Impact of Opinion-Oriented Conversation Design}
\label{sec:corona_result}
Since one of our major focuses for our socialbot was on incorporating opinion-driven conversations, we now present some empirical results quantifying the importance of opinion sharing in chat.

Over a period of three days in early April, we conducted an A/B test of our fact-oriented coronavirus opening and our opinion-oriented coronavirus opening, keeping the remainder of our system stable. Half of our users received the fact-oriented opening whereas the other received the opinion-oriented opening. Table \ref{tab:corona} shows the results of our A/B test, suggesting that opinion-oriented conversation is better received among users, even when the broader conversation domain is the same.

\begin{table}[htbp]
\centering
\begin{tabular}{ l || c }
\textbf{Type} & \textbf{Average Rating} \\
\hline \hline
Fact    & 3.59 \\
Opinion & \textbf{3.73}*  
\end{tabular}
\caption{Results of A/B test for fact-based and opinion-based coronavirus openings. (*) denotes a statistically significant result, $p < 0.10$.}
\label{tab:corona}
\end{table}

\subsection{Impact of Personal Conversation Design}

In addition to our analysis of the opinion-oriented conversation, we investigate the impact of conversation focusing specifically on personal experiences and information. We calculated the average rating of conversations that contained the \textit{Life} component and the average rating of conversations that contained a conversational style that did not focus on personal experience sharing: \textit{Movies}, \textit{Music}, \textit{Virtual Reality}, and \textit{Sports}. The results of this analysis appear in Table \ref{tab:life}. Note that, as described in Section \ref{sec:flow}, we distinguish personal conversation from opinion-oriented conversation that does not contain personalized elements. The four above components whose ratings comprise the "Info and Opinions" category in Table  \ref{tab:life} have an opinion-oriented focus, but do not focus on personal information. We exclude other topics from our analysis, such as the pets component, since it has some elements of personal information sharing as well as many fact-oriented elements. Additionally, we removed conversations with 3 or less turns to reduce noise from uncooperative users. We observed a promising indication that this personal-experience-focused component was more preferable to our users.

\begin{table}[htbp!]
\centering
\begin{tabular}{ l || c}
\textbf{Type} & \textbf{Average Rating} \\
\hline \hline
Info and Opinions & 3.79 \\
Personal Experience  & \textbf{3.85}*
\end{tabular}
\vspace{0.5em}
\caption{ The average rating of conversations stratified by the types of components they contain. (*) denotes a statistically significant result, $p < 0.10$.}
\label{tab:life}
\end{table}

\subsection{Impact of Current Events on Conversation Satisfaction}

\begin{figure}[!htbp]
\begin{subfigure}{.5\textwidth}
  \centering
  \includegraphics[width=0.8\columnwidth]{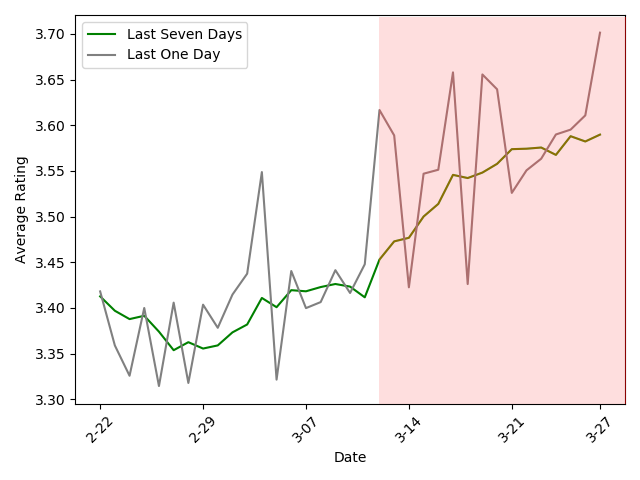}
    \caption{The period when COVID-19 was the opening.}
    \label{fig:corona}
\end{subfigure}
\begin{subfigure}{.5\textwidth}
  \centering
  \includegraphics[width=0.8\columnwidth]{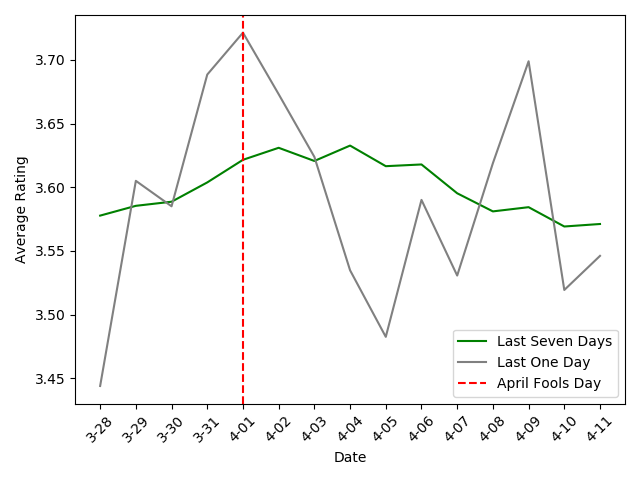}
    \caption{The day that April Fools Day was the opening.}
    \label{fig:aprilfools}
\end{subfigure}

\begin{subfigure}{.5\textwidth}
  \centering
    \includegraphics[width=0.8\columnwidth]{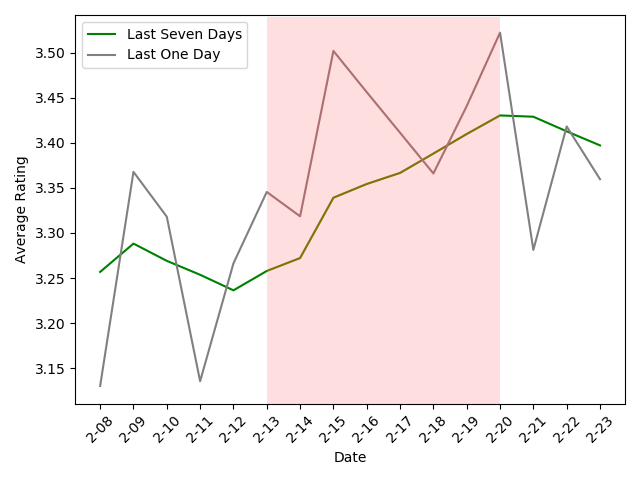}
    \caption{The period when Valentine's Day was the opening.}
    \label{fig:valentines}
\end{subfigure}
\begin{subfigure}{.5\textwidth}
  \centering
    \includegraphics[width=0.8\columnwidth]{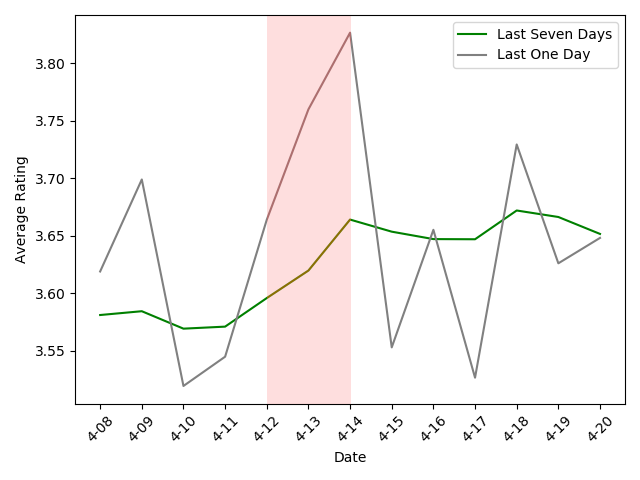}
    \caption{The period when Easter was the opening.}
    \label{fig:easter}
\end{subfigure}
\caption{The average daily and last seven day user ratings for the time periods surrounding our current-event-focused openings. Single day openings are indicated by a red dashed line and multi-day openings are denoted by the red highlighted section in each graph.}
\label{fig:currentevents}
\end{figure}

We focused some of our experimental efforts on testing the importance of portraying awareness of current events to the user. To this end, during the course of Quarterfinals and Semifinals, we implemented various alternative Openings that specifically targeted holidays and current events. For Valentine's Day, April Fool's Day, and Easter, we deployed these alternative Openings instead of our traditional Opening. Although this is not a perfectly rigorous scientific test due to other confounding changes occurring in our system at the same time, Figure \ref{fig:currentevents} presents an indication that discussing current holidays has a positive impact on the user's satisfaction and impression of our socialbot. All time periods in which the Holiday openings were deployed display a positive trend in average user ratings. Similar to the Holiday openings, our COVID-19 opening shown in Figure \ref{fig:currentevents}(a) also displays a positive effect on the user ratings. Unlike the Holiday components, whose underlying current event is more transient, COVID-19 has become a dominating event in current times and, in line with this, the positive impact of our COVID-19 Opening seems to have been sustained over time. 

\section{Conclusion}

Emora is a state-transition-based dialogue manager which combines curated informational resources, highly expressive natural language templates, powerful intent classification, and ontology resources to interact with human users in a compelling manner. The strong performance of Emora in the tail end of this year's Alexa Prize Competition indicates the potential gain that can be achieved through exploring the integration of opinion-and-experience-oriented dialogue strategies in chat-oriented conversations. Our experiences in this competition provide strong motivation to pursue this direction in future work.

In particular, to advance in the direction of experience-oriented chatbots, our emphasis for future work is on automating the development of a socialbot's personal characterization (including their experiences, preferences, and opinions) and designing dynamic adaptation of such characterization over time influenced by current world trends and events. We also will seek to derive automatic methods for connecting distinct experience-oriented topics with one another. In addition, another integral component of open-domain chat that remains is the addition of common-sense knowledge and reasoning into a chatbot's dialogue strategy. There are many challenges within this area, most notably the difficulty in acquiring such commonsense data relevant to conversational exchanges since, by nature, it is generally implicit and assumed by speakers within conversational contexts. Although we were unable to address this in the current year's competition, our future work will aim to address such challenges.

\bibliography{refs}
\bibliographystyle{acl}

\end{document}